%% file: arxivScenGE.tex
\documentclass{article}
\usepackage[preprint]{neurips_2025}
\usepackage[utf8]{inputenc} 
\usepackage[T1]{fontenc}    
\usepackage[colorlinks,linkcolor=red,anchorcolor=blue,citecolor=green]{hyperref}       
\usepackage{url}
\usepackage{booktabs}       
\usepackage{amsfonts}       
\usepackage{nicefrac}       
\usepackage{microtype}      
\usepackage{amsmath}
\usepackage{graphicx}
\usepackage{enumitem}
\usepackage{multirow}
\usepackage{makecell}
\usepackage{wrapfig}
\usepackage{pifont}         
\usepackage{subcaption}
\usepackage{nicematrix}
\usepackage{algorithm}
\usepackage{algorithmic}
\numberwithin{equation}{section}
\input{math_commands.tex}
\newcommand{\Tref}[1]{Tab.~\ref{#1}}
\newcommand{\Eref}[1]{Eq.~(\ref{#1})}
\newcommand{\Fref}[1]{Fig.~\ref{#1}}
\newcommand{\Sref}[1]{Sec.~\ref{#1}}

\newcommand{\eg}{\textit{e}.\textit{g}.}

\newcommand{\wrt}{\textit{w}.\textit{r}.\textit{t}\space}
\newcommand{\toolns}{\textsc{ScenGE}}
\newcommand{\tool}{\toolns\space}
\newcommand{\msgns}{Meta-Scenario Generation}
\newcommand{\msg}{\msgns\space}
\newcommand{\csens}{Complex Scenario Evolution}
\newcommand{\cse}{\csens\space}
\newcommand{\acgns}{Adversarial Collaborator Graph}
\newcommand{\acg}{\acgns\space}

\newcommand*{\affaddr}[1]{#1} 
\newcommand*{\affmark}[1][*]{\textsuperscript{#1}}

\definecolor{knowledge_prior}{HTML}{2ecc71}
\definecolor{LLM}{HTML}{e74c3c}
\definecolor{collaborator_graph}{HTML}{3498db}
\definecolor{temporal_modeling}{HTML}{9b59b6}
\definecolor{perturbation_ratio}{HTML}{f1c40f}
\definecolor{loss}{HTML}{8B4513}

\title{Adversarial Generation and Collaborative Evolution of Safety-Critical Scenarios for Autonomous Vehicles}

\author{
    \textbf{Jiangfan Liu\affmark[1], Yongkang Guo\affmark[1], Fangzhi Zhong\affmark[1], Tianyuan Zhang\affmark[1], Zonglei Jing\affmark[1],}
    \\
    \textbf{Siyuan Liang\affmark[2], Jiakai Wang\affmark[3], Mingchuan Zhang\affmark[4], Aishan Liu\affmark[1]\thanks{Corresponding Author}, Xianglong Liu\affmark[1,3]}
    \\
    \affaddr{\affmark[1]{Beihang University}}
    \affaddr{\affmark[2]{Nanyang Technological University}} \\
    \affaddr{\affmark[3]{Zhongguancun Laboratory}}
    \affaddr{\affmark[4]{Henan University of Science and Technology}} 
}

\begin{document}

\maketitle

\input{Sections/0_Abstract}
\input{Sections/1_Introduction}
\input{Sections/2_RelatedWork}

\input{Sections/3_Method}
\input{Sections/4_Experiment}
\input{Sections/5_Conclusion}

\bibliographystyle{plain}
\bibliography{arxivScenGE}

\end{document}

%% file: math_commands.tex
\usepackage{amsmath,amsfonts,bm}

\def\eqref#1{equation~\ref{#1}}

\def\1{\bm{1}}

\DeclareMathAlphabet{\mathsfit}{\encodingdefault}{\sfdefault}{m}{sl}
\SetMathAlphabet{\mathsfit}{bold}{\encodingdefault}{\sfdefault}{bx}{n}

%% file: Sections/0_Abstract.tex
\begin{abstract} 

The generation of safety-critical scenarios in simulation has become increasingly crucial for safety evaluation in autonomous vehicles (AV) prior to road deployment in society. However, current approaches largely rely on predefined threat patterns or rule-based strategies, which limit their ability to expose diverse and unforeseen failure modes. To overcome these, we propose \toolns, a framework that can generate plentiful safety-critical scenarios by reasoning novel adversarial cases and then amplifying them with complex traffic flows. Given a simple prompt of a benign scene, it first performs \textit{\msgns}, where a large language model (LLM), grounded in structured driving knowledge (\eg, traffic regulations, real-world accident records), infers an adversarial agent whose behavior poses a threat that is both plausible and deliberately challenging. This meta-scenario is then specified in executable code for precise in-simulator control. Subsequently, \textit{\csens} uses background vehicles to amplify the core threat introduced by Meta-Scenario. It builds an adversarial collaborator graph to identify key agent trajectories for optimization. These perturbations are designed to simultaneously reduce the ego vehicle's maneuvering space and create critical occlusions. Extensive experiments conducted on multiple reinforcement learning (RL) based AV models show that \tool uncovers more severe collision cases (+31.96\%) on average than SoTA baselines. Additionally, our \tool can be applied to large model based AV systems and deployed on different simulators; we further observe that adversarial training on our scenarios improves the model robustness. Finally, we validate our framework through real-world vehicle tests and human evaluation, confirming that the generated scenarios are both plausible and critical. We hope our paper can build up a critical step towards building public trust and ensuring their safe deployment. Our codes can be found at \url{https://scenge.github.io}.
\end{abstract}

%% file: Sections/1_Introduction.tex
\section{Introduction}
\label{sec: Introduction}

As autonomous vehicles (AVs) \cite{hu2023_uniad, shao2024lmdrive, ma2024dolphins,li2022bevformer} approach widespread deployment, ensuring their safety and reliability to earn public trust has become critical~\cite{liang2022parallel,liang2020efficient,liang2022large,liu2025natural}. However, this endeavor is challenging due to the scarcity of real-world data on rare yet critical incidents. \textit{Simulation-based testing} \cite{cai2025text2scenario, lu2024multimodal} offers a controlled, reusable, and cost-effective way of evaluating behavior under various conditions, particularly safety-critical scenarios that probe their safety capacity. Current scenario generation approaches, which largely rely on predefined threat templates \cite{zhang2024chatscene, xu2025diffscene, wang2024drivedreamer} or rule-based strategies \cite{ding2020learning, wang2021advsim, scenariorunner, zhang2022adversarial}, struggle to reveal the full spectrum of safety flaws. Due to these weak \emph{risk exposure} abilities, AVs retain undiscovered vulnerabilities from inadequate validation. This challenge is closely related to the field of adversarial machine learning. In this domain, research focuses on exposing the vulnerabilities of deep neural networks through various attack methodologies. Such work also involves evaluating model robustness and developing more resilient architectures \cite{wang2021dual, liu2019perceptual, liu2020bias, zhang2021interpreting, tang2021robustart, liang2023badclip, liang2024object, liu2021training, liu2020spatiotemporal, liu2023x, liu2022harnessing, liu2023exploring, guo2023towards, liu2023towards, liang2025revisiting}.

\begin{wrapfigure}{r}{0.5\textwidth}
    \includegraphics[width=\linewidth]{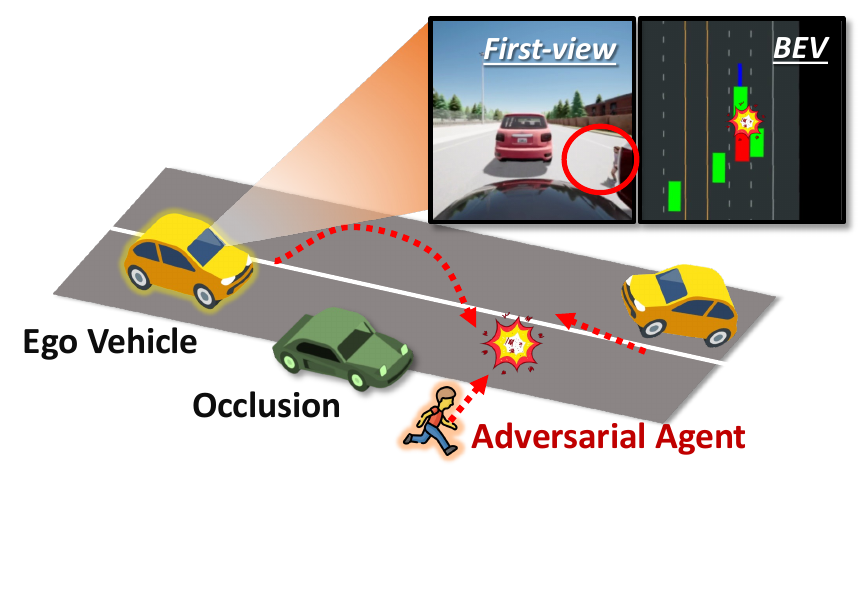}
    \vspace{-30pt}
    \caption{The illustration of the safety-critical scenarios generated by our \toolns: a pedestrian emerging from behind a parked truck, with the threat amplified by a background vehicle obstructing the line-of-sight to the adversary.}
    \label{fig: 1}
\end{wrapfigure}

To address these limitations, we propose \toolns, a two-stage framework that exposes safety vulnerabilities in AV by performing adversarial threat generation and collaborative trajectory evolution. Our first stage, \textit{\msgns}, uses a LLM to creatively generate a core adversarial threat from a benign text prompt. To ensure this threat is both plausible and challenging, a retrieval augmented generation RAG \cite{wu2024retrieval} framework grounds the LLM's reasoning in a structured knowledge base of traffic regulations, driver qualification standards, and realistic pre-crash scenarios. The generated meta-scenario is then expressed as executable Scenic code \cite{fremont2019scenic, fremont2022scenic}, allowing for diverse and scalable instantiation in the CARLA simulator \cite{Dosovitskiy17}. However, the threat created by a single adversarial agent is often predictable and insufficient to create a truly critical dilemma for the AV. Our second stage, \textit{\csens}, therefore crafts more complex threats by coordinating the surrounding background traffic. It first builds an adversarial collaborator graph to identify the most influential vehicles. The trajectories of these key agents are then carefully optimized. Rather than causing direct collisions or simple chaos, these optimizations intensify interaction complexity by strategically limiting the ego's escape paths and obstructing its line-of-sight, ultimately increasing the likelihood of a critical incident.

Extensive experiments on multiple RL-based AV models demonstrate that \tool uncovers more severe collision cases (+31.96\%) on average than state-of-the-art baselines. We also confirm its broad generalizability: the scenarios effectively challenge advanced VLM models like LMDrive, are transferable to the MetaDrive simulator. Beyond testing, we demonstrate the practical value of our data through adversarial training. Models trained on our scenarios exhibit substantially improved robustness. This improvement is validated on real-world nuScenes data, where the enhanced models make demonstrably safer decisions. To bridge the sim-to-real gap, we further validate our approach with real-world vehicle tests and human driver surveys, which confirm our generated scenarios represent plausible and critical real-world risks. Our main \textbf{contributions} are:

\begin{itemize}
    \item We propose \toolns, a two-stage framework that generates safety-critical scenarios by seamlessly combining knowledge-grounded LLM reasoning with multi-agent trajectory optimization.

    \item We introduce two core components: \msgns, which generates richly detailed meta-scenarios by grounding an LLM's reasoning in knowledge priors; \csens, which enhances the resulting threats by optimizing the trajectory of key background vehicles identified via an \acgns.

    \item Extensive experiments conducted on RL-based AV models show the effectiveness of \tool (+31.96\% collision rate on average) compared to state-of-the-art baselines.
\end{itemize}

%% file: Sections/2_RelatedWork.tex
\section{Related Work}
\label{sec: Related Work}

\textbf{Simulation-Based Testing for AV.} Simulation-based testing, facilitated by platforms such as CARLA \cite{Dosovitskiy17}, MetaDrive \cite{li2021metadrive}, and LimSim \cite{wen2023limsim}, offers a cost-effective and controlled environment for evaluating AVs under diverse driving conditions. A key advantage is the ability to replicate rare yet critical scenarios that are impractical or hazardous to test in the real world. This capability for controlled, repeatable testing is critical for systematically identifying performance vulnerabilities and ultimately validating the safety of AVs.

\textbf{Safety-Critical Scenario Generation.} The generation of safety-critical scenarios is crucial for evaluating AV safety, with existing approaches including generative models that learn from real-world data \cite{ding2020learning, tan2021scenegen, suo2021trafficsim, rempe2022generating, feng2023trafficgen}, optimization-based methods that synthesize scenarios via tailored objectives \cite{chen2021adversarial, wang2021advsim, cao2022advdo, xiang2023v2xp}, and semantic-driven methods that incorporate high-level context \cite{gao2023magicdrive, zhong2023language, wang2024drivedreamer, wang2024driving, Zheng2025OccWorld}. More recent works leverage language models for generation from text or logs \cite{li2024chatsumo, zhao2024chat2scenario}, or enhance coverage through semantic replay and trajectory compression \cite{tian2024llm, feng2023dense}.

However, these approaches primarily focus on single-agent rule violations or replaying observed patterns, thus failing to reveal novel, compound failure modes. Even recent LLM-based methods, while improving semantic coverage, still lack fine-grained control over multi-agent interactions. These \textbf{limitations} motivate \toolns, our framework that addresses semantic novelty and emergent risk through structured generation and collaborative perturbation.

%% file: Sections/3_Method.tex
\section{\tool Approach}
\label{sec: Approach}

\begin{figure*}
    \centering
    \includegraphics[width=\linewidth]{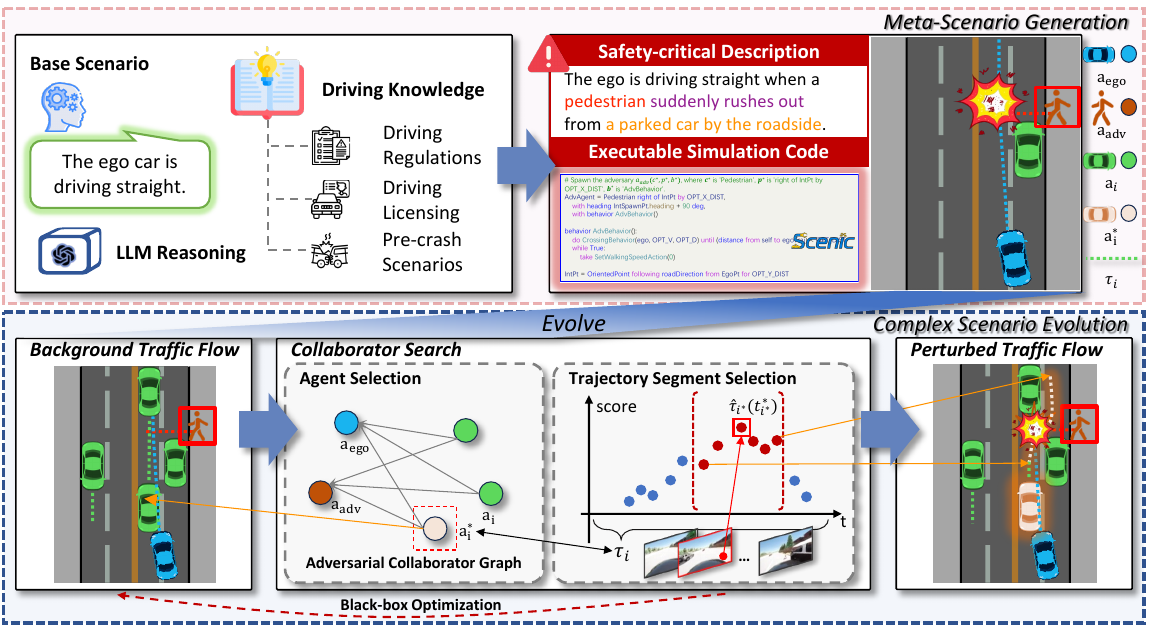}
    \caption{\emph{Framework overview}. Given a simple description of base scenario, \tool first performs \msgns to generate an executable meta-scenario, grounded in violations of established driving knowledge prior. \cse then perturbs the trajectories of key agents within the complex traffic environment to maximize adversarial impact.}
    \label{fig: framework}
\end{figure*}

\tool is designed to target AV system failures stemming from explicit rule violations and subtle multi-agent interactions. An overview is illustrated in \Fref{fig: framework}.

\subsection{Problem Definition}

Let $\mathcal{S}_{\mathrm{meta}} = \{\mathbf{a_{\mathrm{ego}}}, \mathbf{a_{\mathrm{adv}}} \mid \mathrm{R}, \mathrm{L}\}$ denote the \textbf{meta scenario}, which includes an ego vehicle $\mathbf{a}_{\mathrm{ego}}$ and a single adversarial agent $\mathbf{a}_{\mathrm{adv}}$ operating within an environmental context defined by road type $\mathrm{R}$ and traffic light state $\mathrm{L}$. The adversarial agent is further specified by semantic properties $(\mathrm{c}, \mathrm{p}, \mathrm{b})$, denoting its type, position, and behavior. To construct such scenarios, we start from a simple prompt of a benign scene $\Phi_{\mathrm{base}}$, augmented by a fixed instruction prompt $\Phi_{\mathrm{inst}}$ to induce safety-violating behavior. A retrieval function $f_R$ selects relevant entries from a knowledge base $\mathbb{D}$, and the resulting context is used by an LLM $f_{\mathrm{LLM}}$ to produce semantic descriptions $\langle {\Phi}_\mathrm{c}, {\Phi}_\mathrm{p}, {\Phi}_\mathrm{b}, {\Phi}_\mathrm{R}, {\Phi}_\mathrm{L} \rangle$ for adversarial agent and environment properties. These are subsequently parsed into structured values $(\mathrm{c}, \mathrm{p}, \mathrm{b}, \mathrm{R}, \mathrm{L})$ and instantiated to define $\mathcal{S}_{\mathrm{meta}}$.

We define the \textbf{adversarial scenario} as $\mathcal{S}_{adv} = \mathcal{S}_{meta} \cup \{\mathbf{a}_1, \dots, \mathbf{a}_N\}$, where $N$ denotes the number of background vehicles. Each background vehicle $\mathbf{a}_i$ follows a trajectory ${\tau}_i = \{(x_t, y_t)\}_{t=0}^{T}$, representing its simulated coordinates over $T$ frames, where $i {\in} \{1, \dots, N\}$. A subset of background vehicles, indexed by $K \subset \{1, \dots, N\}$, is selected for perturbation. Their trajectory segments are optimized to induce collaborative risky behaviors that increase the threat level of $\mathcal{S}_{\mathrm{meta}}$. Specifically, for each selected vehicle $\mathbf{a}_{i^{\star}}$ with $i^{\star} \in K$, we identify a keyframe $t_{i^{\star}}^{\star}$ as the most influential frame, and define the corresponding perturbable segment $\tilde{{\tau}}_{i^{\star}} \subset {\tau}_{i^{\star}}$ as a temporal window centered at this keyframe. These segments are perturbed by optimizing the objective function $\mathcal{L}$, yielding the final adversarial scenario $\mathcal{S}_{adv}$ with optimized segments $\{\tilde{\tau}_{i^{\star}}^{\star}\}_{i^{\star} \in K}$.

\subsection{\msgns}
\label{subsec: Meta-Scenario Generation}

Given a simple prompt $\Phi_{\mathrm{base}}$, which typically specifies a normal traffic situation without threats (\eg, \texttt{the ego car is driving across the corner}), our goal is to construct a meta-scenario $\mathcal{S}_{\mathrm{meta}}$ in which $\mathbf{a}_{\mathrm{adv}}$ introduces a safety-critical threat. The process comprises two main components: \ding{182} constructing a structured driving knowledge base via RAG, and \ding{183} generating an executable scenario description using an LLM informed by that prior.


\textbf{Safety Driving Knowledge Construction.} The knowledge base $\mathbb{D} = \{\mathbf{D}_r, \mathbf{D}_l, \mathbf{D}_c\}$ consists of three components, each representing a distinct aspect of driving knowledge essential for simulating normative and adversarial traffic behavior. \ding{182} $\mathbf{D}_r$ contains 27 driving regulations segmented from official manuals in the USA, Germany, and China, covering behaviors such as lane merging, overtaking, and other maneuvers. \ding{183} $\mathbf{D}_l$ includes 100 standardized driver's license test questions and answers that assess traffic rule knowledge, situational awareness, and safe behavior selection. Together, $\mathbf{D}_r$ and $\mathbf{D}_l$ provide normative behavioral priors intentionally violated to construct safety-critical adversarial behaviors. In contrast, \ding{184} $\mathbf{D}_c$ comprises 14 pre-crash scenarios drawn from taxonomies in the NHTSA Pre-Crash Typology Report \cite{najm2007pre} (\eg, unprotected left turns, red-light violations), providing concrete adversarial patterns for scenario construction. Collectively, these components inform the synthesis of plausible threat scenarios and support the generation of critical adversarial conditions.

\textbf{LLM-Driven Scenario Generation.} Given a simple prompt $\Phi_{\mathrm{base}}$ of a benign scene, the LLM is prompted to generate a detailed, safety-critical scenario by introducing one main adversarial agent $\mathbf{a}_{\mathrm{adv}}$ into the scenario. It infers the agent's properties and the associated environmental context through in-context learning \cite{dong2024survey}. However, simply adopting an LLM may lead to unsafe or unrealistic critical scenarios; thus, we ground the reasoning process in structured driving knowledge. To this end, relevant knowledge is retrieved from the database $\mathbb{D}$ and combined with the instruction prompt $\Phi_{\mathrm{inst}}$ to form the input to the LLM $f_{\mathrm{LLM}}$. The generation process is formalized as:

\begin{equation}
    \langle {\Phi}_\mathrm{c}, {\Phi}_\mathrm{p}, {\Phi}_\mathrm{b}, {\Phi}_\mathrm{R}, {\Phi}_\mathrm{L} \rangle = f_{\mathrm{LLM}}\left(
    \mathbf{a}_{\mathrm{ego}}, \Phi_{\mathrm{base}}, f_R(\mathbb{D}, \Phi_{\mathrm{base}}) \mid \Phi_{\mathrm{inst}}
\right),
\end{equation}

\noindent where each $\Phi_{*}$ represents a text description of a scenario element, including the adversarial agent's properties and environmental context. The instruction prompt $\Phi_{\mathrm{inst}}$ explicitly guides the model to generate rule-violating yet plausible actions, grounded in retrieved safety knowledge. Although expressed in textual form, the generation is controlled through few-shot prompting and slot-based templates, ensuring the outputs remain structured and scenario-compatible.

The generated descriptions are then parsed into structured values $(\mathrm{c}, \mathrm{p}, \mathrm{b}, \mathrm{R}, \mathrm{L})$ and populated into a predefined Scenic template. This template encodes scenario-level semantics while enforcing syntactic and physical constraints, bridging language-driven generation and executable simulation. The resulting program is run in the simulator to instantiate $\mathcal{S}_{\mathrm{meta}}$.

\subsection{\csens}

Building on the generated meta-scenario, \cse enhances its complexity by introducing background vehicles $\{\mathbf{a}_1, \dots, \mathbf{a}_N\}$ with collaborative risky trajectories. To that end, their interactions with $\mathbf{a}_{\mathrm{adv}}$ and $\mathbf{a}_{\mathrm{ego}}$ are adjusted to create a more challenging scenario for the AV. This process comprises two main components: \ding{182} \textit{Collaborator Search}, which identifies the background vehicles that can most amplify the adversarial nature of the scenario, and \ding{183} \textit{Trajectory Perturbation}, which adjusts the selected vehicles to maximize the adversarial impact.

\textbf{Collaborator Search.} To identify influential background vehicles, we construct an \acg $\bm{G}$, where each node corresponds to an agent in the scenario, and the edges reflect directional behavioral relevance, particularly emphasizing the impact of background vehicles on the ego vehicle and adversarial agent. This graph is derived from a frame-wise attention matrix ${\bm{M}}_{\mathrm{a}}$ that models trajectory-level dependencies using ego and adversarial trajectories as queries and background trajectories as keys. Specifically:

\begin{equation}
{\bm{M}}_{\mathrm{a}} = \frac{({\tau}_{\mathrm{ego}}, {\tau}_{\mathrm{adv}}) \cdot {({\tau}_1, \dots, {\tau}_N)}^\top}{\sqrt{d}} + {\bm{M}}_{\mathrm{m}} + \log {\bm{M}}_{\mathrm{d}},
\end{equation}

\noindent where $d$ is the dimension of $\tau$, ${\bm{M}}_{\mathrm{m}}$ enforces causality by preventing attention to future frames, and ${\bm{M}}_{\mathrm{d}}$ introduces a temporal decay bias to emphasize recent interactions.

Based on ${\bm{M}}_{\mathrm{a}}$, we perform \textit{Collaborator Search} in two stages. First, we aggregate attention scores across frames to estimate each background vehicle's relevance to the ego vehicle and the adversarial agent. This process identifies the $\mathrm{Top\text{-}k}$ most influential vehicles, which form the collaborator set $K$. Then, for each $i^{\star} \in K$, we locate the keyframe $t_{i^{\star}}^{\star}$ receiving the highest attention score for vehicle $\mathbf{a}_{i^{\star}}$, and extract a local temporal window $\tilde{{\tau}}_{i^{\star}}$ centered at keyframe as its perturbable trajectory segment. These segments serve as the input to the subsequent trajectory perturbation module.

\textbf{Trajectory Perturbation.} We optimize the perturbable segments $\{\tilde{{\tau}}_{i^{\star}}\}_{i^{\star} \in K}$ of selected collaborators (indexed by $K$) to maximize the adversarial impact on the ego vehicle. This is formulated as the following objective:

\begin{equation}
\{\tilde{{\tau}}_{i^{\star}}^{\star}\}_{i^{\star} \in K} = \underset{\{\tilde{{\tau}}_{i^{\star}}\}_{i^{\star} \in K}}{\arg\max}~\mathcal{L}(\tilde{{\tau}}_{\mathrm{ego}}, \tilde{{\tau}}_{\mathrm{adv}}, \{\tilde{{\tau}}_{i^{\star}}\}_{i^{\star} \in K}),
\end{equation}

The optimization follows an iterative, gradient-based procedure. Specifically, for each perturbable segment, we compute the gradient of $\mathcal{L}$ \wrt the trajectory coordinates and update them in the direction that increases the loss. Each update step uses a small, fixed step size and is projected back to the feasible space to ensure realism. The process continues until convergence or a predefined number of steps is reached.

\begin{align}
\label{eq: loss}
\mathcal{L} &= 
    \lambda_1 \underbrace{\| \tilde{\tau}_{i^{\star}} - \tilde{\tau}_{\text{ego}} \|_2}_{\mathcal{L}_{\text{ego}}}
    + 
    \lambda_2 \underbrace{\| (\tilde{\tau}_{i^{\star}} - \tilde{\tau}_{\text{ego}}) \times (\tilde{\tau}_{\text{adv}} - \tilde{\tau}_{\text{ego}}) \|_{\perp}}_{\mathcal{L}_{\text{occ}}}
    + 
    \lambda_3 \underbrace{\|\Delta^2 \tilde{\tau}_{i^{\star}}\|_2^2}_{\mathcal{L}_{\text{smooth}}}.
\end{align}

The objective function $\mathcal{L}$ comprises three components, as shown in \Eref{eq: loss}. \ding{182} $\mathcal{L}_{\text{ego}}$ minimizes the Euclidean distance between the perturbed background trajectory $\tilde{\tau}_{i^{\star}}$ and the ego trajectory $\tilde{\tau}_{\text{ego}}$ within a temporal window. \ding{183} $\mathcal{L}_{\text{smooth}}$ penalizes second-order differences $\Delta^2 \tilde{\tau}_{i^{\star}}$ to reduce abrupt motion changes. \ding{184} $\mathcal{L}_{\text{occ}}$ minimizes the normalized perpendicular distance via a 2D cross product, promoting alignment along the ego–adversary line-of-sight. From a behavioral modeling perspective, $\mathcal{L}_{\text{ego}}$ encourages spatial proximity to induce planning hesitation, $\mathcal{L}_{\text{smooth}}$ ensures kinematic feasibility via smoothness constraints, collectively balancing adversarial strength with physical plausibility, and $\mathcal{L}_{\text{occ}}$ amplifies perceptual ambiguity through occlusion. Finally, the optimized perturbations $\{\tilde{\tau}_{i^{\star}}^{\star}\}_{i^{\star} \in K}$ replace the corresponding segments of the original trajectories, yielding the final adversarial scenario $\mathcal{S}_{adv}$, which poses a significant threat to the ego vehicle's safe driving.

\subsection{Overall Generation Workflow} 

Our workflow begins with the \msg module. This module takes a benign text description as input. First, an LLM generates a detailed textual description of a safety-critical threat. This generation process is grounded in a structured driving knowledge base. The LLM then translates this textual description into a parameterized Scenic script. Finally, this script is instantiated within the CARLA simulator to create the executable meta-scenario.

Next, we process the background traffic for the meta-scenario. We use CARLA's Traffic Manager to generate a flow of background vehicles and record their baseline trajectories. The \cse module then analyzes these trajectories offline. It builds an \acg to identify the most influential background vehicles to act as collaborators. The module selects critical segments of their trajectories and optimizes them to maximize the overall threat. This optimization aims to restrict the ego vehicle's maneuvering space and create critical occlusions. For the final evaluation, we replay the complete scenario in a closed-loop simulation. The optimized background vehicles execute their new adversarial trajectories as scripted events, forcing the ego vehicle to react to the coordinated, high-risk situation.

%% file: Sections/4_Experiment.tex
\section{Experiment and Evaluation}
\label{sec: Experiment and Evaluation}

\begin{table}
    \caption{\textbf{Evaluation of adversarial scenario generation methods across CR, and OS  metrics.} Performance is assessed on eight base scenarios in CARLA, averaged across PPO, SAC, and TD3 models. Best results are highlighted in bold. Higher CR and lower OS values, indicate better adversarial effectiveness.}
    \label{tab: 1}
    \centering
    \resizebox{\linewidth}{!}{
        \begin{tabular}{@{}c|c|cccccccc|c@{}}
            \toprule
            \multirow{3}{*}{\textbf{Metric}} & \multirow{3}{*}{\textbf{Algo.}} & \multicolumn{8}{c|}{\textbf{Base Traffic Scenarios}} &                                                                                                                                       \\
                                             &                                 & \scriptsize{\makecell{Straight                                                                                                                                                               \\ Obstacle}} & \scriptsize{\makecell{Turning \\ Obstacle}} & \scriptsize{\makecell{Lane \\ Changing}}  & \scriptsize{\makecell{Vehicle \\ Passing}} & \scriptsize{\makecell{Red-light \\ Running}} & \scriptsize{\makecell{Unprotected \\ Left-turn}} & \scriptsize{\makecell{Right-\\ turn}} & \scriptsize{\makecell{Crossing \\ Negotiation}} & \multirow{-2}{*}{\textbf{Avg.}} \\
            \midrule
            \multirow{5}{*}{CR \textcolor{red}{$\uparrow$}}
                                             & LC                              & 0.241                                                & 0.159          & 0.736          & 0.792          & 0.317          & 0.325          & 0.321          & 0.313          & 0.401          \\
                                             & AS                              & 0.451                                                & 0.399          & 0.726          & 0.832          & 0.177          & 0.335          & 0.115          & 0.303          & 0.417          \\
                                             & CS                              & 0.391                                                & 0.679          & 0.756          & 0.812          & 0.237          & 0.325          & 0.411          & 0.333          & 0.493          \\
                                             & AT                              & 0.441                                                & 0.379          & 0.646          & 0.782          & 0.317          & 0.315          & 0.321          & 0.353          & 0.440          \\
                                             & ChatScene                       & 0.750                                                & 0.647          & 0.660          & \textbf{0.907} & \textbf{0.833} & 0.620          & 0.743          & 0.850          & 0.751          \\
            \cline{2-11}
                                             & \textbf{Ours}                   & \textbf{0.860}                                       & \textbf{0.773} & \textbf{0.837} & 0.897          & 0.823          & \textbf{0.747} & \textbf{0.763} & \textbf{0.863} & \textbf{0.820} \\
            \midrule
            \multirow{5}{*}{OS \textcolor{blue}{$\downarrow$}}
                                             & LC                              & 0.789                                                & 0.816          & 0.566          & 0.530          & 0.799          & 0.790          & 0.692          & 0.717          & 0.712          \\
                                             & AS                              & 0.694                                                & 0.687          & 0.561          & 0.506          & 0.866          & 0.775          & 0.841          & 0.721          & 0.706          \\
                                             & CS                              & 0.726                                                & 0.552          & 0.549          & 0.513          & 0.839          & 0.787          & 0.649          & 0.708          & 0.665          \\
                                             & AT                              & 0.696                                                & 0.706          & 0.599          & 0.528          & 0.805          & 0.795          & 0.689          & 0.698          & 0.690          \\
                                             & ChatScene                       & 0.559                                                & 0.572          & 0.607          & 0.472          & 0.544          & 0.656          & 0.511          & \textbf{0.459} & 0.548          \\
            \cline{2-11}
                                             & \textbf{Ours}                   & \textbf{0.503}                                       & \textbf{0.526} & \textbf{0.504} & \textbf{0.457} & \textbf{0.507} & \textbf{0.519} & \textbf{0.498} & 0.477          & \textbf{0.499} \\
            \bottomrule
        \end{tabular}
    }
\end{table}

\subsection{Experimental Setup}
\label{subsec: Experimental Setup}

\textbf{Simulation environment and benchmark.} We utilise the CARLA simulator \cite{Dosovitskiy17}, an open-source and highly customizable urban driving simulator, to create a closed-loop simulation environment. We adopt SafeBench \cite{xu2022safebench} as the benchmarking framework, which supports diverse RL-based AV agents and standardized evaluation. Following \cite{zhang2024chatscene}, we use 8 base traffic scenarios (\eg, \texttt{Straight Obstacle, Lane Changing}) curated from the NHTSA Pre-Crash Typology Report \cite{najm2007pre}, each containing 10 diverse driving routes. For each route, 10 adversarial scenarios are generated, resulting in 800 challenging scenarios for evaluation and comparison per method.

\textbf{AV algorithms.} Following \cite{zhang2024chatscene}, we mainly employ 3 representative RL-based AV algorithms as testing agents, including Proximal Policy Optimization (PPO) \cite{schulman2017proximal}, Soft Actor-Critic (SAC) \cite{haarnoja18b}, and Twin Delayed Deep Deterministic Policy Gradient (TD3) \cite{fujimoto18a}.

\textbf{Compared baselines.} We compare our \tool with several existing scenario generation methods, including Learning-to-Collide (LC) \cite{ding2020learning}, AdvSim (AS) \cite{wang2021advsim}, Carla Scenario Generator (CS) \cite{scenariorunner}, Adversarial Trajectory Optimization (AT) \cite{zhang2022adversarial}, and ChatScene \cite{zhang2024chatscene}. For fair comparisons, each method is applied on the same 8 base scenarios and routes to generate 800 challenging scenarios under consistent generation logic and evaluation settings.

\textbf{Metrics.} Following SafeBench \cite{xu2022safebench}, we adopt a set of key metrics to evaluate AV performance in generated scenarios. Two core indicators are used: the \textbf{collision rate} (CR \textcolor{red}{$\uparrow$}) measures the frequency of collisions and reflects safety risk, and the \textbf{overall score} (OS \textcolor{blue}{$\downarrow$}) aggregates system-level performance. In addition, we evaluate three additional dimensions: the \textbf{safety level} (\textit{frequency of running red lights} (RR \textcolor{red}{$\uparrow$}), \textit{frequency of running stop signs} (SS \textcolor{red}{$\uparrow$}), and \textit{average distance driven out of road} (OR \textcolor{red}{$\uparrow$}), the \textbf{functionality level} (\textit{route following stability} (RF \textcolor{blue}{$\downarrow$}), \textit{average percentage of route completion} (Comp \textcolor{blue}{$\downarrow$}), and \textit{average time spent to complete the route} (TS \textcolor{red}{$\uparrow$})), and the \textbf{etiquette level} (\textit{average acceleration} (ACC \textcolor{red}{$\uparrow$}), \textit{average yaw velocity} (YV \textcolor{red}{$\uparrow$}), and \textit{frequency of lane invasion} (LI \textcolor{red}{$\uparrow$}). Higher (\textcolor{red}{$\uparrow$}) values indicate worse performance, while \textcolor{blue}{$\downarrow$} indicates the contrary.

\textbf{Implementation details.} All experiments were conducted on a server with an Intel(R) Core(TM) i9-14900K CPU and two NVIDIA GeForce RTX 4090 GPUs with 24GB memory. The LLM used for \msgns is qwq-32b \cite{qwq32b}, the reasoning model from the Qwen series. In the \cse module, we construct 10 background vehicles and perturb the trajectories of 4 selected ones, each over 60\% of their trajectory. The 4 perturbed vehicles are selected based on the highest attention relevance to ego and adversarial agents. The 60\% perturbation window is centered around each vehicle’s most relevant keyframe. We set $\gamma$ in the decay matrix to $0.8$. In the loss calculation, we set $\lambda_1 = 0.3$, $\lambda_2 = 0.2$, and $\lambda_3 = 0.5$.

\subsection{Main Results}
\label{subsec: Main Results}

Our main results, presented in \Tref{tab: 1} and \Tref{tab: 2}, compare \tool against baseline methods across eight base scenarios. To ensure a robust and generalizable evaluation, all metrics are averaged over three distinct RL agents, as detailed in \Sref{subsec: Experimental Setup}. This approach validates that our scenarios pose a universal challenge to a range of modern driving policies, rather than merely exploiting agent-specific flaws. The tables offer complementary views: \Tref{tab: 1} details the CR and OS for each individual scenario, whereas \Tref{tab: 2} assesses the overall impact on AV behavior from the three key aspects of \textit{safety and risk exposure}, \textit{functionality under stress}, and \textit{driving etiquette}, with all results averaged across the scenarios.

\begin{wraptable}{r}{0.6\textwidth}
    \caption{\textbf{Aggregated evaluation results across safety, functionality, and etiquette dimensions.} Each value represents the average over three RL-based AV agents and eight base scenarios.}
    \label{tab: 2}
    \resizebox{0.6\columnwidth}{!}{
    \begin{tabular}{@{}c|ccc|ccc|ccc@{}}
        \toprule
            \multirow{2}{*}{\textbf{Algo.}} & \multicolumn{3}{c|}{\textbf{Safety Level}} & \multicolumn{3}{c|}{\textbf{Functionality Level}} & \multicolumn{3}{c}{\textbf{Etiquette Level}} \\
            & RR \textcolor{red}{$\uparrow$} & SS \textcolor{red}{$\uparrow$} & OR \textcolor{red}{$\uparrow$} & RF \textcolor{blue}{$\downarrow$} & Comp \textcolor{blue}{$\downarrow$} & TS \textcolor{red}{$\uparrow$} & ACC \textcolor{red}{$\uparrow$} & YV \textcolor{red}{$\uparrow$} & LI \textcolor{red}{$\uparrow$} \\
        \midrule
            LC            & \textbf{0.325} & 0.165 & 0.039 & 0.884 & 0.807 & 0.224 & 0.225 & 0.231 & 0.087 \\
            AS            & 0.299 & 0.167 & 0.032 & 0.901 & 0.821 & 0.269 & 0.217 & 0.233 & 0.102 \\
            CS            & 0.312 & \textbf{0.168} & \textbf{0.043} & 0.880 & 0.817 & 0.252 & 0.229 & 0.235 & 0.106 \\
            AT            & 0.311 & 0.167 & 0.035 & 0.883 & 0.802 & \textbf{0.287} & 0.233 & 0.236 & 0.112 \\
            ChatScene     & 0.228 &	0.145 & 0.018 & 0.890 &	0.571 &	0.074 &	0.281 &	0.225 &	0.064 \\
        \midrule
            \textbf{Ours} & 0.231 & 0.125 & 0.009 & \textbf{0.838} & \textbf{0.472} & 0.124 & \textbf{0.402} & \textbf{0.359} & \textbf{0.179} \\
        \bottomrule
        \end{tabular}
    }
\end{wraptable}

\textbf{Safety and Risk Exposure.} As shown in \Tref{tab: 1}, \tool achieves the highest average CR. Notably, this high collision rate is not achieved by forcing simplistic rule violations. In fact, the scores for RR, SS, and OR are not the highest, as seen in \Tref{tab: 2}. This outcome is a direct result of our design philosophy. We avoid creating simplistic, predictable setups that rely on obvious rule violations. Instead, \tool focuses on generating high-pressure situations from complex multi-agent interactions. These plausible scenarios challenge the predictive and planning capabilities under pressure. They force the vehicle to navigate moments where no single, simple rule applies. Consequently, the resulting collisions expose more profound and subtle vulnerabilities in the AV's core logic, rather than surface-level failures in rule compliance.

\textbf{Functionality Challenges.} As shown in \Tref{tab: 1}, \tool achieves the lowest average OS. Additionally, \Tref{tab: 2} shows a \textbf{4.96\%} drop in RF and a \textbf{29.16\%} reduction in Comp. The moderate TS is caused by early collisions, which demonstrates that \tool induces rapid and decisive failures by persistently disrupting the AV's planning.

\begin{wrapfigure}{r}{0.5\textwidth}
    \includegraphics[width=\linewidth]{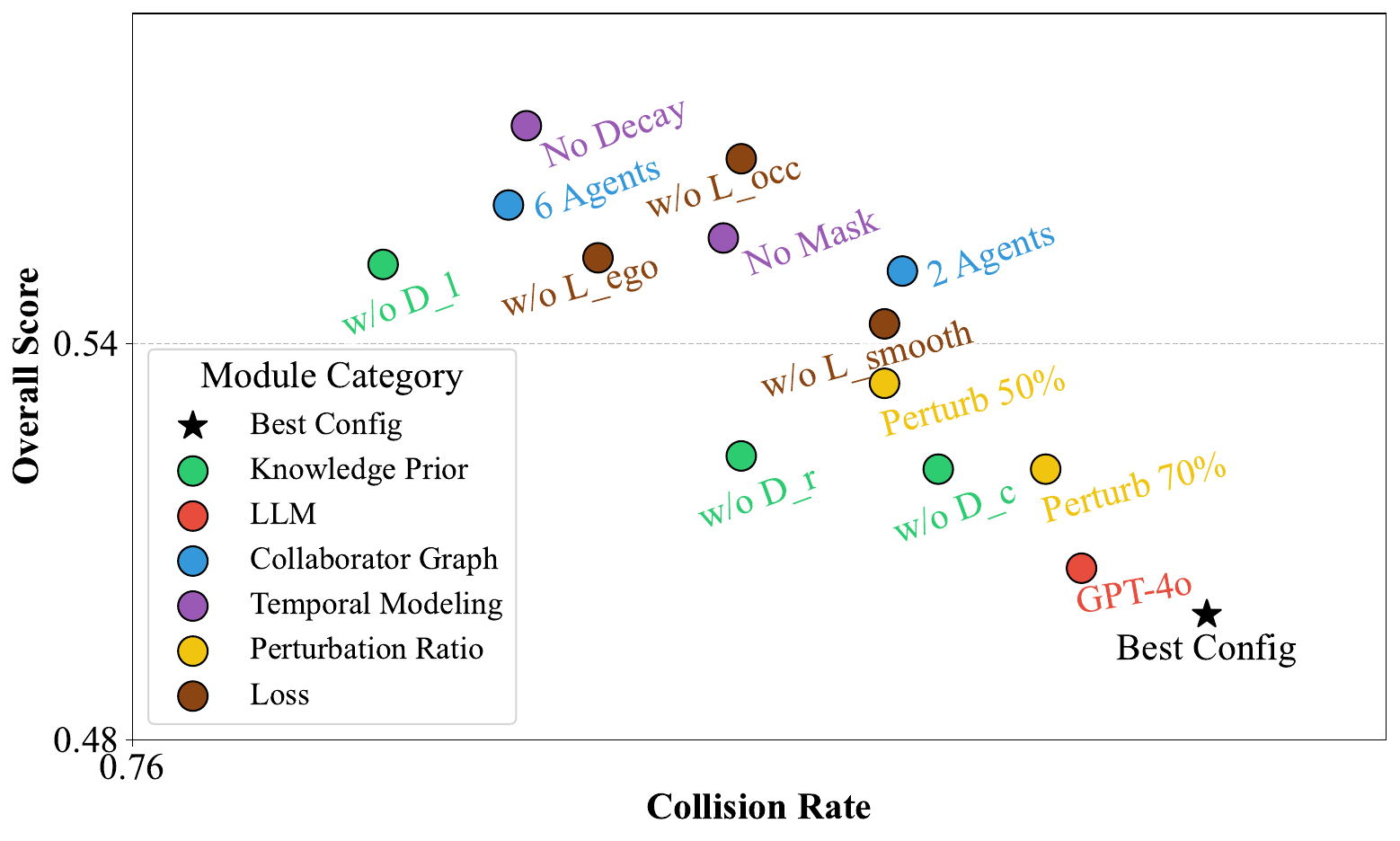}
    \caption{Scatter plot of {\textbf{CR} \textcolor{red}{$\uparrow$}} vs. {\textbf{OS} \textcolor{blue}{$\downarrow$}} across ablation settings. Each color denotes an ablation type, and each point represents a specific variant. Points closer to the bottom-right indicate stronger adversarial effects.} 
    \label{fig: ablation}
\end{wrapfigure}

\textbf{Driving Etiquette.} As shown in \Tref{tab: 2}, \tool increases ACC, YV, and LI by \textbf{16.5\%}, \textbf{12.7\%}, and \textbf{8.48\%} respectively. These results suggest that \tool causes AV to behave less smoothly and more erratically in ways that remain socially plausible. Introducing temporally coordinated perturbations across multiple agents disrupts fine-grained control and social driving compliance, revealing limitations that simpler, single-agent or rule-based methods fail to expose.

\subsection{Ablation Studies}
\label{subsec: Ablation Studies}

We perform ablation experiments by selectively disabling key modules and observing the effect on performance. Otherwise specified, this part keeps the same setting as the main experiment. \Fref{fig: ablation} reports the results under different settings. 

\ding{182} {\color{knowledge_prior} Knowledge Prior.} Removing $\mathbf{D}_r$ yields CR 79.4\% and OS 52.3\%, reflecting its role in guiding rule-focused violations. Removing $\mathbf{D}_l$ gives CR 77.4\% and OS 55.2\%, showing its effect on enhancing logical consistency in behavior. Removing $\mathbf{D}_c$ results in CR 80.5\% and OS 52.1\%, confirming its importance in producing realistic and high-risk scenarios. 

\ding{183} {\color{LLM} LLM.} Both GPT-4o and qwq-32b demonstrate strong capabilities, yielding similar CRs of 81.3\% and 82.0\%, respectively. This suggests that the performance is not critically sensitive to the specific choice of a powerful LLM. 

\ding{184} {\color{collaborator_graph} Collaborator Graph.} We ablate the number of perturbed agents by selecting 2, 4, and 6 collaborators. Perturbing 4 agents performs best with CR 82\% and OS 49.9\%, balancing adversarial strength and scenario plausibility. Interestingly, perturbing 6 agents (CR 78.1\%) is less effective than perturbing 2 (CR 80.3\%). We hypothesize that this occurs because an excessive number of agents transforms our intended coordinated, precise threat into easily avoidable chaos. Their mutual interference and abnormal behavior likely prompt the AV to adopt a conservative policy. This hypothesis is also supported by OS, where the 6-agent setting yields a higher score (56.1\%) compared to the 2-agent setting (55.1\%). 

\ding{185} {\color{temporal_modeling} Temporal Modeling.} The full setting (with both mask and decay) yields the best result with CR 82\% and OS 49.9\%. Removing the temporal mask reduces temporal causality in collaborator selection, leads to CR 79.3\% and OS 55.6\%, while removing the temporal decay results in CR 78.2\% and OS 57.3\%. These results highlight the complementary role of both components in capturing temporally coherent influence. 

\ding{186} {\color{perturbation_ratio} Perturbation Ratio.} We compare three perturbation ratios centered around the selected keyframe: 50\%, 60\%, and 70\%. Perturbing 60\% of the segment achieves the best result with CR 82\% and OS 49.9\%. 50\% leads to CR 80.2\% and OS 53.4\%, indicating insufficient behavioral deviation, while 70\% causes CR 81.1\% and OS 52.1\% due to over-modification and reduced plausibility. These results shows moderate ratio balances realism and adversarial effect. 

\ding{187} {\color{loss} Loss.} We ablate each component in $\mathcal{L}$ to assess its contribution. Removing $\mathcal{L}_{\text{ego}}$ leads to CR 78.6\% and OS 55.3\%, reflecting reduced collision targeting. Excluding $\mathcal{L}_{\text{smooth}}$ yields CR 80.2\% and OS 54.3\%, with trajectories becoming visibly unstable. Removing $\mathcal{L}_{\text{occ}}$ results in CR 79.4\% and OS 56.8\%, indicating weaker alignment between adversary and ego. The full loss yields the best trade-off, and ablating any term consistently reduces CR and increases OS.

\subsection{Generalization Ability Analysis}
\label{subsec: Evaluation on VLM AV Models}

\begin{wraptable}{r}{0.55\textwidth}
    \caption{Performance of LMDrive under generated adversarial scenarios across eight base traffic tasks.}
    \label{tab: vlm}  
    \resizebox{0.55\columnwidth}{!}{
    \begin{tabular}{@{}c|ccc@{}}
        \toprule
            {\textbf{Algo.}} & \textbf{Benign Scenario} & \textbf{Meta-Scenario} & \textbf{Adversarial Scenario} \\
        \midrule
            \textbf{RC} & 92.2 \textsubscript{$_\text{\textcolor{blue}{$\pm$ 2.9}}$} & 92.9 \textsubscript{$_\text{\textcolor{blue}{$\pm$ 2.7}}$} & 89.9 \textsubscript{$_\text{\textcolor{blue}{$\pm$ 5.1}}$} \\
            \textbf{IS} & 0.97 \textsubscript{$_\text{\textcolor{blue}{$\pm$ 0.01}}$} & 0.9 \textsubscript{$_\text{\textcolor{blue}{$\pm$ 0.05}}$} & 0.89 \textsubscript{$_\text{\textcolor{blue}{$\pm$ 0.04}}$} \\
            \textbf{DS} & 87.7 \textsubscript{$_\text{\textcolor{blue}{$\pm$ 2.4}}$} & 83.7 \textsubscript{$_\text{\textcolor{blue}{$\pm$ 4.7}}$} & 80.4 \textsubscript{$_\text{\textcolor{blue}{$\pm$ 5.5}}$} \\
        \bottomrule
    \end{tabular}
    }
\end{wraptable}

We study the generalizability of \toolns: \ding{182} effectiveness on other AV models; and \ding{183} applicability on other simulators.

\textbf{Model Generalization.} Beyond RL-based AV models, we further evaluate our generated scenarios on LMDrive \cite{shao2024lmdrive}, a large vision-language model for AV deployed on the CARLA Leaderboard \cite{leaderboard}. LMDrive navigates by following natural language instructions sequentially, using the multi-view camera and Lidar perception for scene understanding and planning. To accommodate its instruction-driven execution mode, we redesign the test routes into multi-instruction sequences. Evaluation follows LMDrive's original metrics: Route Completion (RC), Infraction Score (IS), and Driving Score (DS). We evaluate LMDrive under three increasingly challenging settings: \ding{182} ego-only benign routes as a baseline, \ding{183} meta-scenarios with a single adversarial agent, and \ding{184} full adversarial scenarios generated by our framework, including the perturbed background vehicle. As shown in \Tref{tab: vlm}, LMDrive's performance drops from \textbf{87.7} DS in the benign case to \textbf{83.7} in meta-scenarios and further to \textbf{80.4} under full adversarial conditions. These results demonstrate that our generated scenarios significantly stress LMDrive's planning capability, especially under rare or occluded interactions. 

\begin{wrapfigure}{r}{0.6\textwidth}
    \includegraphics[width=\linewidth]{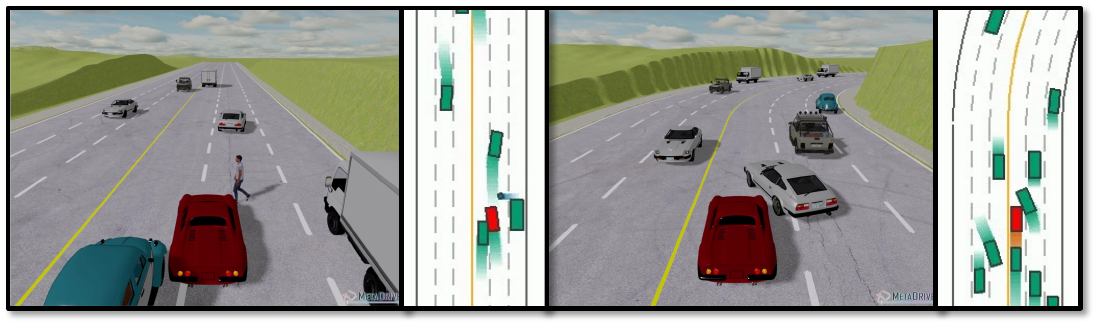}
    \caption{Applicability of \tool on MetaDrive simulator.}
    \label{fig: metadrive}
\end{wrapfigure}

\textbf{Simulator Generalization.} We validated the generalizability of \tool in MetaDrive \cite{li2021metadrive} simulator, a lightweight platform for RL research. Despite substantial differences in rendering and physics, \toolns's modular design facilitated a successful port. As shown in \Fref{fig: metadrive}, our framework successfully generated effective safety-critical scenarios that induced high CR in MetaDrive. This result demonstrates the broad applicability of our approach.

\subsection{Training on the Generated Scenarios}
\label{subsec: Adversarial Training on the Generated Scenarios}

This section evaluates the utility of our \tool in enhancing AV robustness by adversarial training on the generated scenarios. Specifically, we evaluate the robustified RL agent on both simulated data and real-world data.

\textbf{Robustness evaluation in simulation.} We adversarially train the SAC-based ego vehicle across eight base traffic scenarios using scenes from the first eight routes per scenario, and evaluate on unseen scenes from the remaining two routes. The training process uses 500 epochs with a learning rate of 0.0001. As shown in \Tref{tab: adv_training}, adversarial training with \toolns-generated scenarios yields the best results among methods, reducing the CR by \textbf{3.1\%} while increasing the OS by \textbf{94.7\%}. These results demonstrate that our method generates scenarios missing from standard AV training. Adversarial training on these data remedies the AV vulnerabilities, leading to a significant improvement in robustness.

\begin{wraptable}{r}{0.6\textwidth}
    \caption{Evaluation of adversarially trained ego vehicle.}
    \label{tab: adv_training}  
    \resizebox{0.6\columnwidth}{!}{
    \begin{tabular}{@{}c|ccccc|c@{}}
        \toprule
            {\textbf{Metric}} & LC & AS & CS & AT & ChatScene & \textbf{Ours}\\
        \midrule
            CR \textcolor{red}{$\uparrow$}            & 0.210 & 0.216 & 0.176 & 0.135 & 0.043 & \textbf{0.031}\\
            OS \textcolor{blue}{$\downarrow$}            & 0.813 & 0.806 & 0.825 & 0.864 & 0.905 & \textbf{0.947} \\
        \bottomrule
    \end{tabular}
    }
\end{wraptable}

\textbf{Robustness evaluation on real-world data.} We further evaluated our adversarially trained model on real-world data from the nuScenes dataset \cite{caesar2020nuscenes} to address the visual domain gap between simulation and reality. In particular, we first manually select 140 scenario segments (2700 images) with latent risks, such as a pedestrian standing by the roadside; subsequently, we evaluate both the original RL model and the enhanced model on these real-world image sequences. The enhanced model achieved a \textbf{21.7\% lower} Euclidean distance to the ground truth trajectory than the original model, indicating safer and more stable decisions. This result demonstrates that the robustness gained from our scenarios transfers effectively to real-world perception data.

\subsection{Real-world Experiments}

\begin{wrapfigure}{r}{0.6\textwidth}
    \includegraphics[width=\linewidth]{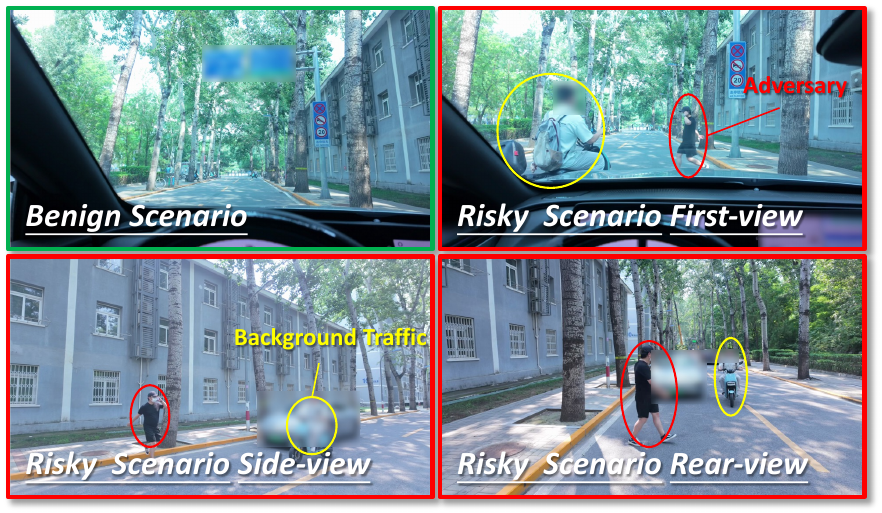}
    \caption{Real-World Experiments.}
    \label{fig: realworld}
\end{wrapfigure}

Here, we conducted real-world experiments in a closed road, where we arranged a layout that is similar to our generated scenario and tested a real-world vehicle (\Fref{fig: realworld}). Due to commercial confidentiality, the appearance of our test vehicle has been obscured. Under strict safety protocols, we recreated and repeated two challenging scenarios 15 times each: a pedestrian suddenly emerging from an occlusion, and an unprotected left-turn challenged by an accelerating scooter. Despite safety measures designed to lower the scenarios' difficulty (\eg, limiting vehicle speed), the AV exhibited critical failures. In \textbf{73.3\%} of the pedestrian tests, the vehicle failed to react to the pedestrian's emergence in a timely manner. Similarly, in \textbf{60\%} of the left-turn tests, it failed to alter its trajectory to avoid the conflicting scooter, exposing a decisive vulnerability. These results provide definitive, physical-world evidence that our generated scenarios identify physical risks for autonomous systems, not just simulation artifacts.

\subsection{Discussion and Analysis}

\textbf{Computational Cost.} We compare the per-scenario generation cost of \tool against the baselines in \Sref{subsec: Experimental Setup}. When generating in batches (\eg, 5 Scenic scripts, each instantiating 20 variations), our framework averages approximately 0.024 GPU hours per scenario. In contrast, AS requires 0.06 GPU hours, AT needs 0.13 GPU hours, and LC takes 0.17 GPU hours. Our approach thus reduces the required generation time by \textbf{60\%} compared to AS, \textbf{81.5\%} compared to AT, and \textbf{85.9\%} compared to LC. Furthermore, methods like CS and ChatScene are bottlenecked by non-scalable expert effort. This demonstrates \toolns's superior efficiency and scalability for large-scale AV testing.

\textbf{Human Evaluation.} We conducted a human study with 30 licensed drivers to assess the real-world plausibility and perceived danger of our generated scenarios. In the study, participants viewed 40 unique videos and rated them using a 5-point scale. The survey covered all eight base scenarios used in our experiments, with five distinct variations selected for each. To mitigate order effects and learning biases, the videos were presented in a fully randomized sequence. An analysis of 1,200 responses revealed high average scores for both plausibility (\textbf{4.765} out of 5) and perceived risk (\textbf{4.934} out of 5). This confirms that human drivers consider the scenarios to be both highly realistic and critically dangerous.

\begin{figure*}
    \centering
    \begin{subfigure}{0.49\linewidth}
        \includegraphics[width=\linewidth]{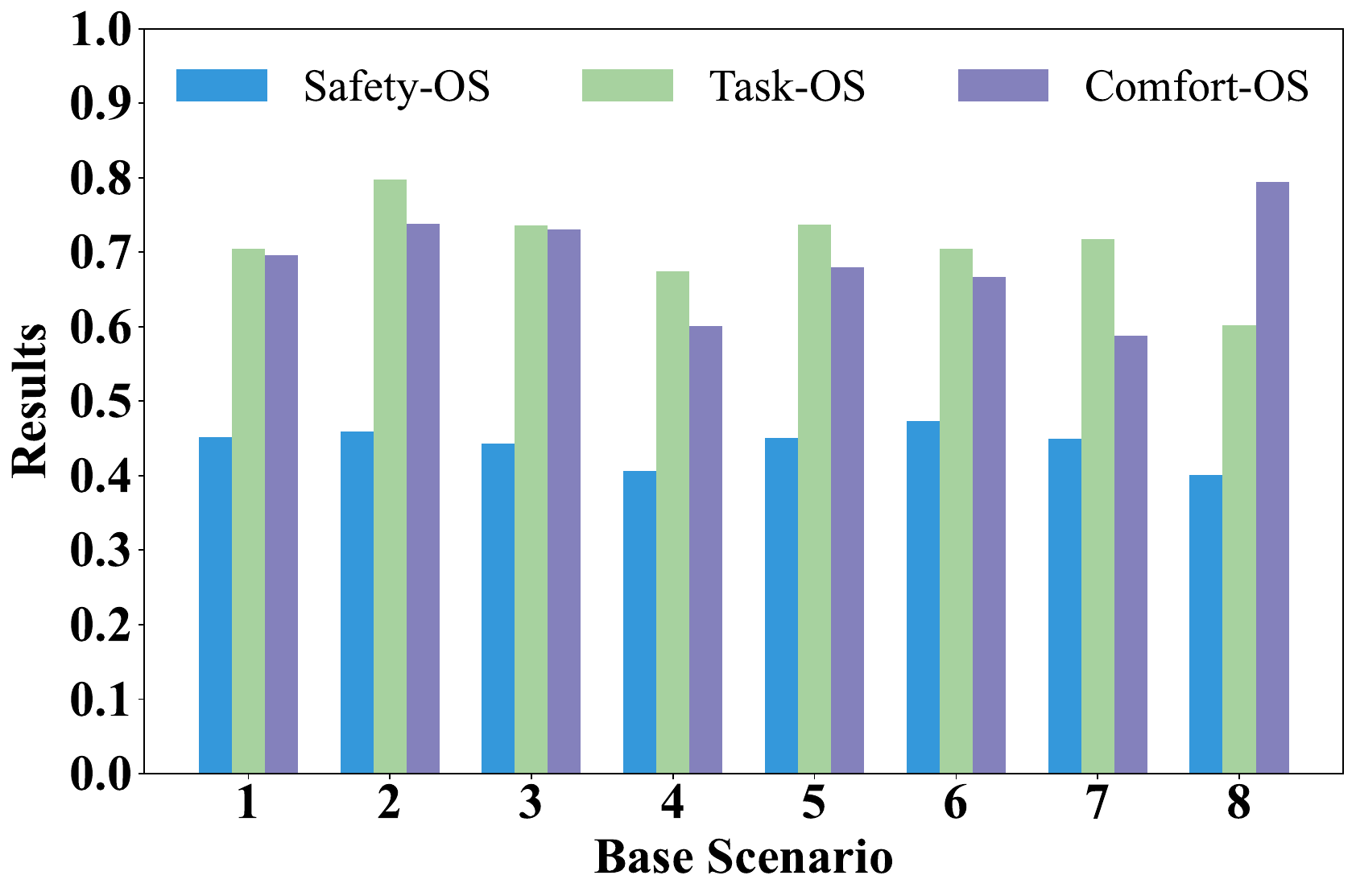}
        \caption{Metric scores of the SAC-based ego vehicle across 8 adversarial scenarios.}
        \label{fig: hist}
    \end{subfigure}
    \hfill
    \begin{subfigure}{0.49\linewidth}
        \includegraphics[width=\linewidth]{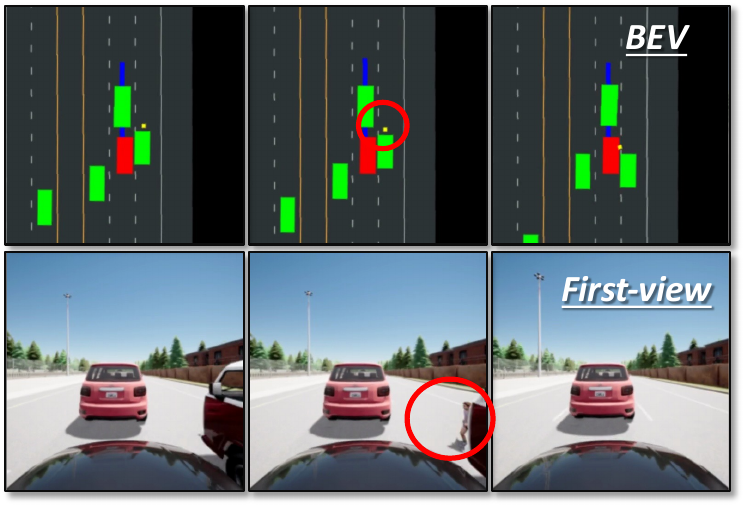}
        \caption{Representative frames  highlight how the agents are jointly leading to a collision.}
        \label{fig: case}
    \end{subfigure}
    \caption{Model performance and case visualization.}
    \label{fig: vis}
\end{figure*}

\textbf{Case Study.} As shown in \Fref{fig: hist}, we evaluate the SAC-based ego vehicle across three composite metrics: safety, task completion, and comfort. These metrics summarize AV behavior by aggregating relevant low-level indicators. We select Scenario 1 for detailed analysis because its scores lie near the median across all three dimensions, making it a representative failure case. In \Fref{fig: case}, we show some frames from the final adversarial scenario. This scenario involves a pedestrian emerging from an occlusion. The threat is amplified by background vehicles: one occupies the adjacent lane to restrict maneuvering space (a behavior promoted by $\mathcal{L}_{\text{ego}}$), while another obstructs the line-of-sight to the adversary (guided by $\mathcal{L}_{\text{occ}}$). These compounded interactions prevent the ego vehicle from executing a safe evasive action. This case effectively illustrates how \tool uses subtle, coordinated behaviors of background traffic to expose critical AV vulnerabilities.

%% file: Sections/5_Conclusion.tex
\section{Conclusion and Future Work}
\label{sec: Conclusion}

In this paper, we introduce \toolns, a two-stage framework for generating safety-critical scenarios to expose vulnerabilities in AV. From a benign scene description, \tool introduces \textit{\msgns} and \textit{\csens} to generate scenarios that are more likely to cause failures. Experiments on multiple RL-based AV models show that \tool reveals more severe collision cases. \textbf{Future work:} A primary direction is to collaborate with automotive manufacturers, integrating our \tool to evaluate the safety of their proprietary, real-world AV algorithms and systems.

